\documentclass[runningheads]{llncs}
\usepackage{graphicx}
\usepackage{hyperref}

\usepackage{comment}
\usepackage{adjustbox}
\usepackage{multirow}
\usepackage{booktabs}
\usepackage{csquotes}
\usepackage{amsmath}
\usepackage{graphicx}
\graphicspath{ {./images/} }
\usepackage{svg}
\usepackage{threeparttable}

\begin{document}
\title{Effective Context Selection in LLM-based Leaderboard Generation: An Empirical Study}
\titlerunning{Effective Context Selection in LLM-based Leaderboard Generation}
%
\author{Salomon Kabongo\inst{1}\orcidID{0000-0002-0021-9729} \and
Jennifer D'Souza\inst{2,3}\orcidID{0000-0002-6616-9509} \and
S\"oren Auer\inst{3}\orcidID{0000-0002-0698-2864}}

\authorrunning{S. Kabongo et al.}
%
\institute{L3S Research Center, Leibniz University of Hannover, Hannover, Germany
\email{kabenamualu@l3s.de}\\ \and
TIB Leibniz Information Centre for Science and Technology, Hannover, Germany\\
\email{\{jennifer.dsouza,auer\}@tib.eu}}
\maketitle              
\begin{abstract}
This paper explores the impact of context selection on the efficiency of Large Language Models (LLMs) in generating Artificial Intelligence (AI) research leaderboards, a task defined as the extraction of (Task, Dataset, Metric, Score) quadruples from scholarly articles. By framing this challenge as a text generation objective and employing instruction finetuning with the FLAN-T5 collection, we introduce a novel method that surpasses traditional Natural Language Inference (NLI) approaches in adapting to new developments without a predefined taxonomy. Through experimentation with three distinct context types of varying selectivity and length, our study demonstrates the importance of effective context selection in enhancing LLM accuracy and reducing hallucinations, providing a new pathway for the reliable and efficient generation of AI leaderboards. This contribution not only advances the state of the art in leaderboard generation but also sheds light on strategies to mitigate common challenges in LLM-based information extraction.
\end{abstract}

\section{Introduction}

The exponential growth of scientific literature~\cite{bornmann2021growth} poses a challenge in tracking AI advancements~\cite{altbach2019too}. Leaderboards, which rank AI models based on performance metrics across various tasks and datasets, serve as essential tools for tracking progress. We refer to the task as SOTA, i.e. state of the art, and it defined as extracting the (Task, Dataset, Metric, Score) quadruples, (T, D, M, S) henceforward. Traditional approaches to SOTA~\cite{axcell,kabongo2021automated,kabongo2023zero} have relied on Natural Language Inference (NLI), which requires a predefined taxonomy and struggles to incorporate new elements. This work introduces a novel approach by framing SOTA as a text generation objective, leveraging Large Language Models (LLMs) and instruction finetuning~\cite{shamsabadi2024large}. Unlike NLI-based methods, this approach adapts to new developments without the need for an exhaustive taxonomy. For finetuning instructions, we rely on the FLAN-T5 collection~\cite{flan-collection}, a comprehensive open-source resource of diverse task-specific instructions, to align our LLMs closely with the domain-specific requirements of the SOTA task. Instruction finetuning~\cite{instructgpt} is instrumental in this context, enabling our LLMs to better align with the domain-specific nuances of leaderboard data, marking a significant departure from previous approaches~\cite{axcell,kabongo2021automated,kabongo2023zero}.
\begin{figure}[!htb]
\centering
\includegraphics[width=8cm]{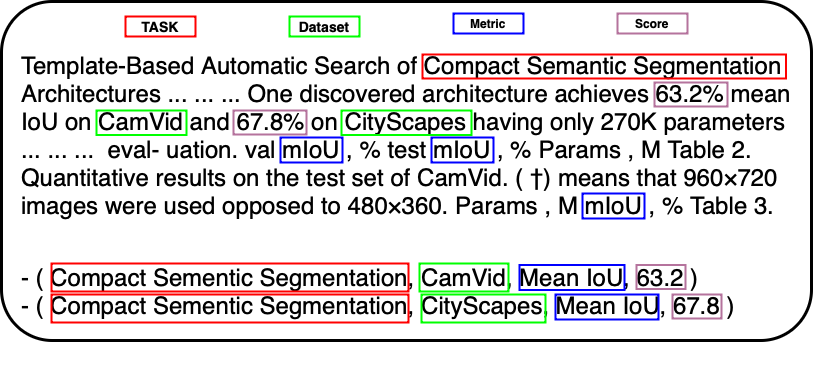}
\caption{Two sets of (Task, Dataset, Metric, Score) tuples reported in an AI paper.}
\label{fig:tdms-example}
\end{figure}

The efficacy of LLMs in information extraction (IE) tasks is heavily influenced by the context provided in the input prompts. As an aside, context for this work implies the discourse text from which information is to be extracted and has no overlap with the well-known notion of in-context learning where one or more successful task examples are given to the LLM to better align its behavior to the given instruction. Specifically, context in this work implies the scholarly article text from which (T, D, M, S) is to be extracted. See ~\autoref{fig:tdms-example}. In this regard, the context plays a dual role: guiding the model towards relevant information extraction~\cite{shamsabadi2024large} and mitigating the risk of hallucination—-a phenomenon where models generate misinformation or irrelevant content. However, longer contexts can serve as distractors~\cite{shi2023large}, leading to decreased model accuracy and increased hallucination rates. Studies~\cite{shi2023large} have shown that models tend to generate more hallucinations when dealing with extensive contexts, as these provide more opportunities for deviation from the task at hand. Addressing this issue requires a careful selection of context~\cite{shi2023large} that balances between providing sufficient guidance and avoiding information overload.

This work addresses these challenges by experimenting with three distinct context types—varying in selectivity and length—to illuminate the critical importance of effective context selection in optimizing LLM performance for the State Of The Art (SOTA) task. Our findings demonstrate the critical importance of effective context selection in enhancing model accuracy and reducing hallucinations. By identifying the optimal context configuration, we provide a roadmap for leveraging LLMs in the efficient generation of AI research leaderboards. This approach not only advances the state of the SOTA task, but also offers insights into mitigating common challenges associated with LLM-based IE tasks. By focusing on the impact of context, this study aims to mitigate the risk of hallucination and distraction, paving the way for more reliable and efficient leaderboard generation.

Our contributions are threefold: Firstly, we thoroughly examine structured summary generation tasks and paper type classification using LLMs. This is encapsulated in our first research question (RQ1): Which context type among DocTAET, DocREC, and DocFULL yields the most accurate performance for generating structured summaries and leaderboard/no-leaderboard classification?

Secondly, building upon RQ1, we turn our focus to the precise extraction of individual elements, addressing our second research question (RQ2): Considering the balance between precision and other performance metrics, which context type presents the best trade-off performance in few-shot and zero-shot settings? This question acknowledges the essential nature of precision in scholarly communication and its broader implications on the fidelity of model outputs.

Lastly, we address the generalizability of LLMs in handling new, unseen tasks, leading to our third research question (RQ3): How does the choice of context affect the generalizability of language models to tasks without prior examples (zero-shot learning)? By comparing these results to our previous approaches, we contribute to the knowledge base regarding LLMs' adaptability and potential for automated research curation.

In addition to these scientific inquiries, we release our \href{https://github.com/Kabongosalomon/LLLM-LeaderboardLLM}{source code} to enable replication and further research. These novel efforts, tested together for the first time, advance the frontier of context-aware performance in language models finetunning and mark a step forward to improve LLM reliability in generating AI research leaderboards.

\begin{table*}[h]
\begin{center}
\begin{threeparttable}
\begin{minipage}{\textwidth}
\begin{tabular*}{\textwidth}{@{\extracolsep{\fill}}l|ccc@{\extracolsep{\fill}}}
\cmidrule{1-4}%
& \multicolumn{3}{@{}c@{}}{\textbf{Our Corpus}} \\\cmidrule{2-4}%
 & Train &Test-Few-shot & Test Zero-shot \\
\midrule
Papers w/ leaderboards & 7,987 & 753& 241  \\
Papers w/o leaderboards &  4,401 & 648 & 548  \\
Total TDM-triples & 415,788 & 34,799 & 14,800  \\
Distinct TDM-triples & 11,998 & 1,917 & 1,267  \\
Distinct \textit{Tasks}       & 1,374 & 322 & 236  \\
Distinct \textit{Datasets}    & 4,816 & 947 & 647  \\
Distinct \textit{Metrics}     & 2,876 & 654 & 412  \\
Avg. no. of TDM per paper & 5.12 & 4.81 & 6.11  \\
Avg. no. of TDMS per paper & 6.95 & 5.81 & 7.86\\
\end{tabular*}
\caption{Our DocREC (Documents Result[s], Experimentation[s] and Conclusion) corpora statistics. The ``papers w/o leaderboard'' refers to papers that do not report leaderboard.}
\label{table:DocREC_Stats}
\end{minipage}
\end{threeparttable}
\end{center}
\end{table*}
\section{Corpus}
\label{sec:corpus}

\noindent{\textbf{Corpus with (T, D, M, S) annotations.}} We created a new corpus as a collection of scholarly papers with their (T, D, M, S) quadruple annotations for evaluating the \textsc{SOTA} task. This dataset is derived from the community-curated (T, D, M, S) annotations for thousands of AI articles available on PwC (CC BY-SA). It articles span Natural Language Processing and Computer Vision domains, among other AI domains such as Robotics, Graphs, Reasoning, etc, thus, being representative for empirical AI research. The specific PwC source download timestamps is \textit{December 09, 2023}. As such the corpus comprised over 8,000 articles with 7,987 used for training and 994 for testing. These articles, originally sourced from arXiv under CC-BY licenses, are available as latex code source, each accompanied by one or more (T, D, M, S) annotations from PwC. While the respective articles' metadata was directly obtained from the PwC data release, the articles collection had to be reconstructed by downloading them from arXiv under CC-BY licenses. Once downloaded, the articles being in `.tex` format needed to undergo pre-processing for tex-to-text conversion so that their contents could be mined. For this, the Pandoc alongside a custom script was applied to extract targeted regions of the paper DocTEAT or DocREC accordingly . Each article's parsed text was then finally annotated with (T, D, M, S) quadruples via distant labeling. Our overall corpus statistics are reported in Table \ref{table:All_Stats}. 

\noindent{\textbf{Corpus with no leaderboards.}} In addition to our base dataset reported in \autoref{table:DocREC_Stats}, we additionally included a set of approximately 4,401 and 648 articles that do not report leaderboards into the train and test sets, respectively. These articles were randomly selected by leveraging the arxiv category feature, then filtering it to papers belonging to domains unrelated to AI/ML/Stats. These articles were annotated with the \textit{unanswerable} label to finetune our language model in recognizing papers without (T,D,M,S) mentions in them.



\noindent{\textbf{The \textsc{SOTA} task objective.}} We phrased the following question to formulate our task objective w.r.t. the (T, D, M, S) extraction target: \textit{What are the values for the following properties to construct a Leaderboard for the model introduced in this article: task, dataset, metric, and score?} In essence, it encapsulates an IE task.

\noindent{\textbf{Instructions for the LLM.}} Instruction tuning~\cite{flan-collection} boosts LLMs by providing specific finetuning instructions, improving adaptability and performance on new tasks~\cite{instructgpt,flan-t5}. This method offers a more efficient approach than traditional unlabeled data methods, allowing for versatile task prompting with single instructions.

In this vein, the ``Flan 2022 Collection''~\cite{flan-collection} was used. The context, in our case, is a selection from specific sections from the full-text of an article where (T, D, M, S) information is most likely shared.

\noindent{\textbf{Context for the LLM.}} In this work, for the first time, we empirically investigate the impact of leveraging different contexts to the LLM. Specifically, we test the following three main types:

\begin{enumerate}
    \item DocTAET: Delineated in prior work, the context to the LLM comprises text selected from the (T)-title, (A)-abstract, (E)-experimental setup, and (T)-tabular information parts of the full-text. It yields an average context length of 493 words, ranging from a minimum of 26 words to a maximum of 7,361 words. These specific selections targeted the areas of the paper where the (T, D, M, S) were most likely to be found.
    \item DocREC: Introduced for the first time in this work, the DocREC context comprises text selected from the sections named (R)-results, (E)-experiments, and (C)-conclusions with allowances for variations in the three names. Complementary but still unique to DocTAET, the DocREC context representation aims to distill the essence of the research findings and conclusions into a succinct format. This context, ended by being much longer than DocTAET, yielded an average length of 1,586 words, with a minimum length of 27 words and a maximum length of 127,689 words
    \item DocFULL: As a last representation and to test the hypothesis that longer contexts which are not selectively tailored to suit the desired task objective tends to distract and thereby hinder the LLM performance, we used the full paper text as context. This approach entailed compiling the LaTeX source code of the document and translating its entirety into a plain text file. DocFULL ended producing the longest contexts compared to DocTAET and DocREC, in an average length of 5,948 words.
\end{enumerate}

\section{Approach}
\label{model}

\begin{figure}[!tb]
\includegraphics[width=\linewidth]{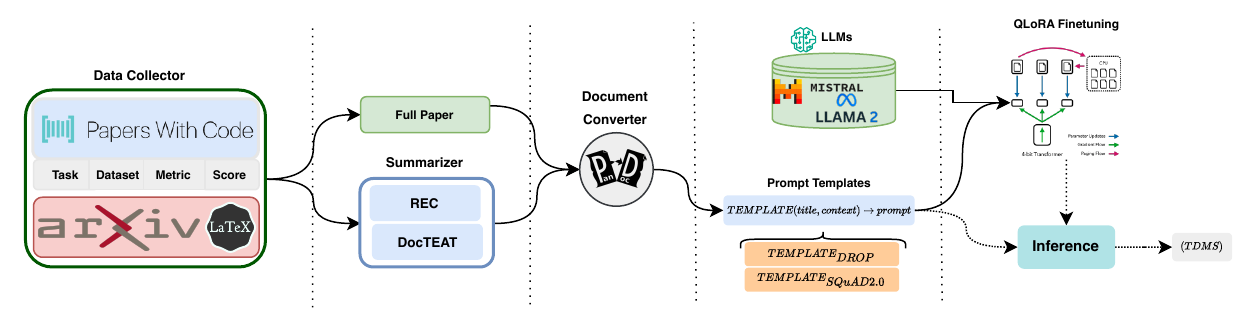}
\caption{Main process overview}
\label{fig:main_process}
\end{figure}

Our approach is single-task instruction-finetuning to address the \textsc{SOTA} task. As such it aims to be an incremental progression of the instruction-tuning paradigm introduced as FLAN (Finetuned Language Net)~\cite{flan-t5,flan-collection}. Equipped with the relevant set of 15 total instructions (8 SQuAD and 7 DROP), we needed to do two things: \textbf{1.} For each instance in the dataset, instantiate the ``Context'' placeholder in the instructions with the DocTAET context feature of a paper and the ``Question'' placeholder with formulated question for the SOTA objective. \textbf{2.} The LLM could then be finetuned with the instruction-instantiated training set. From \autoref{table:DocREC_Stats}, given our training dataset on average had 7,987 (T, D, M, S) papers x 15 instructions x 1 SOTA objective question = 119,805 instruction-instantiated data points to train the LLM. To this, the 4,401 papers without leaderboards x 15 instructions x 1 SOTA objective = 66,015 instruction-instantiated data points were added.

\subsection{Models}


Since Flan-T5~\cite{flan-t5}, there have been several LLMs released in the community that surpassed its performance. In this paper, while we rely on the Flan-T5 instruction finetuning paradigm, we apply the approach on two recent state-of-the-art LLMs on the \href{https://huggingface.co/spaces/HuggingFaceH4/open_llm_leaderboard}{public LLM leaderboard} in a comparative experiment setting.

As the first model, we selected Mistral-7B~\cite{mistral}. This model as the name suggested is a 7-billion-parameter language model optimized for performance and efficiency. It introduces the Grouped-Query Attention (GQA) for rapid inference and reduced memory requirements, and Sliding Window Attention (SWA) for handling long sequences with lower computational costs. The model surpasses existing models in benchmarks, including reasoning, mathematics, and code generation tasks. It also features fine-tuning capabilities for instruction following, achieving superior performance in human and automated benchmarks. \href{https://huggingface.co/mistralai}{Mistral 7B} is designed for real-time applications, supports high throughput, and its architecture enables effective sequence generation with optimized memory usage. The model is released under the Apache 2.0 license, with its \href{https://github.com/mistralai/mistral-src}{source code} on Github, facilitating broad accessibility and application in various tasks.

As the second model, we selected the LLama-2 model~\cite{touvron2023llama}. The Llama-2 model is a collection of LLMs that range from 7 billion to 70 billion parameters, designed for both general and dialogue-specific applications. From the three available \href{https://huggingface.co/meta-llama}{Llama-2 model checkpoints}, i.e. 7B, 13B, and 70B, for comparability with our first select model, i.e. Mistral-7B, we choose the Llama-2 7B model. The Llama-2 models are fine-tuned for enhanced dialogue use cases and exhibit improved performance over existing open-source models in terms of helpfulness and safety, based on benchmarks and human evaluations. The Llama-2 family includes models optimized for different scales and introduces safety and fine-tuning methodologies to advance the responsible development of LLMs.

We employed the QLORA (Quantum-enhanced Learning Optimization for Robust AI) \cite{dettmers2024qlora} framework for fine-tuning our models, leveraging its advanced optimization capabilities to enhance model performance. QLORA has been recognized for its innovative approach to integrating quantum computing principles with machine learning, offering a novel pathway to overcoming traditional optimization challenges.

\section{Evaluations}
\label{eval}

\noindent{\textbf{Experimental setup.}} For training, we had one main experimental setting based on the 15 instructions. As elicited earlier in \autoref{sec:corpus}, i.e. the Corpus section, each of the 15 instruction were instantiated with the 7,987 (T, D, M, S) data instances and the SOTA question resulting in a total of 119,805 instances to instruction finetune both LLAMA2 and Mistral 7B. In this scenario, we hypothesized that this repetition in the data instances across the instructions would cause the resulting model to overfit the training dataset. Thus to control for this, we applied the following experimental setup. Each instruction was instantiated with a random selection of only half the 7,987 (T, D, M, S) data instances resulting in a finetuning dataset of a sizeable 41,340 instances that had leaderboards. The papers w/o leaderboards were also similarly halved. In the test scenario, however, we report per instruction (T, D, M, S) instantiated data results. As shown in \autoref{table:DocREC_Stats}, for a few-shot test set had 994 (T, D, M, S) instances and 648 papers without leaderboards and zero-shot test set had 652 (T, D, M, S) instances and 548 papers without leaderboards, evaluations results are shown for each instruction separately with a total of 2,353 underlying papers representing those with and without leaderboards. Model hyperparameter details are in \autoref{app:hyp}. In terms of compute, all experiments including inference were run on an NVIDIA 3090 GPU. Training took ~20 hours on the 50\% sampled dataset, while inference lasted ~10 minutes for 2,353 test instances.

\begin{table*}[!htb]
  \centering
  \begin{adjustbox}{width=1\textwidth}
  
    \begin{tabular}{|p{2cm}|p{1.1cm}rrrr|rrrrr|l|}
      \toprule
      &        \multicolumn{5}{|c|}{Few-shot} & \multicolumn{5}{c}{Zero-shot} &     \\
      \toprule
    \bf Model & \bf Rouge1 & \bf Rouge2 & \bf RougeL & \bf RougeLsum & \bf \stackbox[c]{General\\-Accuracy} &\
       \bf  Rouge1 & \bf Rouge2 & \bf RougeL & \bf RougeLsum & \bf \stackbox[c]{General\\-Accuracy} & \bf Context \\
      \midrule
    \multirow{3}{*}{Llama-2 7B} & 49.68&10.18 &48.91 &49.02 &83.51 & 68.15& 4.81& 67.59& 67.78& 86.82& Ours \\ \cline{2-12} 
      &49.70 &17.62 &48.81 &48.81 &83.62 & 62.75& 10.88& 62.07& 62.18& 86.22& DocTAET \\ \cline{2-12}
      &5.38 &0.79 &4.96 &5.13 &57.54 &7.55 &0.71 &7.24 &7.35 &37.8 &  FullPaper \\ \hline
    \multirow{3}{*}{Mistral 7B}  & 55.46& 14.11& 54.54& 54.64& 88.44& 72.98& 6.87& 72.42&72.35 & 92.40&  Ours \\ \cline{2-12} 
      & 57.24& 19.67& 56.28& 56.19& \textbf{89.68}&73.54 &12.23 & 73.01&72.95 &\textbf{95.97} &  DocTAET \\ \cline{2-12}
     & 6.73	&	0.77&	6.36&	6.49& 58.45&9.38 	&0.59	&9.11	&9.23	& 39.28& FullPaper \\ \hline
    \end{tabular}
  \end{adjustbox}
  \caption{Evaluation results of LLama-2 and Mistral-7B with output evaluations as a structured summary generation task (reported with ROUGE metrics) as well as binary classification between papers with and without leaderboards (reported as General Accuracy) for each of the 15 instructions from DROP and SQuAD datasets, respectively, with 50\% randomly selected paper instantiations per template.}
  \label{tab:rouge-50percent}
\end{table*}

\noindent{\textbf{Metrics.}} We evaluated our models in two main settings. In the first setting, we treated the \textsc{SOTA} task objective as a structured summarization task. In this setting, we applied standard summarization ROUGE metrics (details in \autoref{app:rouge}). Furthermore, we also tested the models ability to identify papers with leaderboards and those without. This task was simple. For the papers with leaderboards, the model replied with a structured summary and for those it identified as without it replied as ``unanswerable.'' For these evaluations we applied simple accuracy measure. In the second setting, we evaluated the model JSON output in a fine-grained manner w.r.t. each of the inidividual (T, D, M, S) elements and overall for which we reported the results in terms of the standard F1 score and Precision score.

\begin{table*}[!htb]
\centering
\begin{adjustbox}{width=1\textwidth}
	\begin{tabular}{|l|l|ccccc|ccccc|l|} \hline
		&         & \multicolumn{5}{|c|}{Few-shot} & \multicolumn{5}{c}{Zero-shot} &     \\ \toprule
	{\bf Model}	& \bf Mode & \stackbox[c]{\bf Task}  & \bf Dataset & \bf Metric & \stackbox[c]{\bf Score} & \bf Overall & \stackbox[c]{\bf Task} & \bf Dataset & \bf Metric & \stackbox[c]{\bf Score} & \bf Overall & \bf Context \\ \hline
        \multirow{6}{*}{Llama-2 7B} & \textcolor{orange}{Exact} &20.93 & 13.06&13.96 &3.04 &12.75 & 13.97& 6.83& 11.72& 2.61& 8.78& \multirow{2}{*}{Ours} \\ \cline{2-12}
		& \textcolor{blue}{Partial} & 31.37& 22.50&21.99 &3.46 &19.83 & 24.05& 16.6&18.28 &3.10 &15.51 & \\ \cline{2-13}
		\multirow{2}{*}{} & \textcolor{orange}{Exact}  & 29.53 & 16.68 & 20.02 & 1.14 & 16.84 & 21.75 & 11.26 & 16.99 & 0.77 & 12.69 &  \multirow{2}{*}{DocTAET} \\ \cline{2-12}
		& \textcolor{blue}{Partial} & 43.37 & 30.36 & 30.51 & 1.38 & 26.4 & 38.48 & 23.1 & 27.09 & 0.96 & 22.41 & \\ \cline{2-13}
		\multirow{2}{*}{} & \textcolor{orange}{Exact} & 1.59 & 1.36& 0.94& 0.23& 1.03& 2.06 &1.30 &1.52 &0.33 &1.30 & \multirow{2}{*}{FullPaper} \\ \cline{2-12}
		& \textcolor{blue}{Partial}  &2.29 &1.82 &1.68 &0.37 &1.54 & 3.36 &2.49 &2.49 &0.54 &2.22 & \\ \hline
        \multirow{6}{*}{Mistral 7B} & \textcolor{orange}{Exact}  & 26.77 & 15.68& 18.70& 6.36& 16.88&17.99 & 11.80& 15.55& 5.04& 12.60&  \multirow{2}{*}{Ours} \\ \cline{2-12}
		& \textcolor{blue}{Partial}  &39.75 &27.28 &28.49 &7.08 & 25.65& 29.88&21.05 &23.16 & 5.75& 19.96& \\ \cline{2-13}
		\multirow{2}{*}{} & \textcolor{orange}{Exact}  & 33.38 & 18.51& 24.23& 1.87& \textbf{19.50}& 26.99& 14.32& 22.04& 1.20& \textbf{16.14} & \multirow{2}{*}{DocTAET} \\ \cline{2-12}
		& \textcolor{blue}{Partial}  & 46.35& 32.75& 34.16& 2.25& \textbf{28.88}& 44.90& 27.29& 32.23& 1.41& \textbf{26.46}& \\ \cline{2-13}
		\multirow{2}{*}{} & \textcolor{orange}{Exact} & 0.81& 0.57& 0.57& 0.56& 0.63&0.22 &0.33 &0.33 &0.76 & 0.42&  \multirow{2}{*}{FullPaper} \\ \cline{2-12}
		& \textcolor{blue}{Partial} & 1.19& 0.85& 0.81& 0.84& 0.92& 0.56& 0.67& 0.78& 0.87& 0.72&  \\ \hline
	\end{tabular}
\end{adjustbox}
\caption{Evaluation results of Llama-2 and Mistral w.r.t. the individual (Task, Dataset, Metric, Score) elements and Overall in the model JSON generated output in terms of \textbf{F1 score} for each of the 15 instructions from DROP and SQuAD datasets, respectively, with 50\% randomly selected paper instantiations per template.}
\label{tab:f1-50percent}
\end{table*}

\begin{table*}[!htb]
\centering
\begin{adjustbox}{width=1\textwidth}
	\begin{tabular}{|l|l|ccccc|ccccc|l|} \hline
		&         & \multicolumn{5}{|c|}{Few-shot} & \multicolumn{5}{c}{Zero-shot} &     \\ \toprule
	{\bf Model}	& \bf Mode & \stackbox[c]{\bf Task}  & \bf Dataset & \bf Metric & \stackbox[c]{\bf Score} & \bf Overall & \stackbox[c]{\bf Task} & \bf Dataset & \bf Metric & \stackbox[c]{\bf Score} & \bf Overall & \bf Context \\ \hline
        \multirow{6}{*}{Llama-2 7B} & \textcolor{orange}{Exact}  & 34.10& 21.27& 22.74& 4.99& 20.78& 31.89& 15.77& 26.77& 6.06& 20.12&  \multirow{2}{*}{Ours} \\ \cline{2-12}
		& \textcolor{blue}{Partial} & 51.13& 36.66& 35.82& 5.59& 32.32& 54.92& 38.32& 41.73& 7.27& 35.56&  \\ \cline{2-13}
		\multirow{2}{*}{} & \textcolor{orange}{Exact}  & 30.61& 17.29& 20.78& 1.18& 17.46& 24.56& 12.72& 19.19& 0.87& 14.34&  \multirow{2}{*}{DocTAET} \\ \cline{2-12}
		& \textcolor{blue}{Partial}  & 44.96& 31.48& 31.66& 1.43& 27.38& 43.46& 26.09& 30.60& 1.09& 25.31&  \\ \cline{2-13}
		\multirow{2}{*}{} & \textcolor{orange}{Exact}  & 34.69& 29.59& 20.41& 5.10& 22.45& 32.20& 20.34& 23.73& 5.08& 20.34&  \multirow{2}{*}{FullPaper} \\ \cline{2-12}
		& \textcolor{blue}{Partial}  & 50.00& 39.80& 36.73& 8.16& 33.67& 52.54& 38.98& 38.98& 8.47& 34.75&  \\ \hline
        \multirow{6}{*}{Mistral 7B} & \textcolor{orange}{Exact}   & 37.65& 22.15& 26.38& 8.94& 23.78& 35.68& 23.40& 31.02& 9.98& 25.02&  \multirow{2}{*}{Ours} \\ \cline{2-12}
		& \textcolor{blue}{Partial}  &55.90 & 38.52& 40.18& 9.95& 36.14& 59.25& 41.73& 46.20& 11.46& 39.66& \\ \cline{2-13}
		\multirow{2}{*}{} & \textcolor{orange}{Exact}   & 39.48& 21.89& 28.66& 2.21& 23.06& 38.46& 20.41& 31.41& 1.71& 23.00&  \multirow{2}{*}{DocTAET} \\ \cline{2-12}
		& \textcolor{blue}{Partial} & 54.82& 38.73& 40.41& 2.65& 34.15& 64.00& 38.89& 45.94& 2.03& 37.71&  \\ \cline{2-13}
		\multirow{2}{*}{} & \textcolor{orange}{Exact}  & 32.43& 32.43& 32.43& 9.6& \textbf{30.76}&25.00 &37.50 &37.50 &14.00 & \textbf{28.50}&   \multirow{2}{*}{FullPaper} \\ \cline{2-12}
		& \textcolor{blue}{Partial} & 71.43& 48.65& 45.95& 14.52& \textbf{45.13}&62.50 &75.00 &87.50 &21.62 & \textbf{61.66}& \\ \hline
	\end{tabular}
\end{adjustbox}
\caption{Evaluation results of Llama-2 and Mistral w.r.t. the individual (Task, Dataset, Metric, Score) elements and Overall in the model JSON generated output in terms of \textbf{Precision score} for each of the 15 instructions from DROP and SQuAD datasets, respectively, with 50\% randomly selected paper instantiations per template.}
\label{tab:precision-50percent}
\end{table*}

\subsection{Results and Discussion}

This section presents an analysis of the comparative performance of different contexts provided to LLMs, examining their impact on model precision and reliability across various tasks and settings.

\noindent{\textbf{Impact of Context on Structured Summary Generation and Classification (RQ1).}}
Our investigation revealed that the context type profoundly affects the model's ability to generate accurate summaries and classifications. As indicated in \autoref{tab:rouge-50percent}, when equipped with the DocTAET context, both Llama-2 7B and Mistral 7B models consistently outperformed the other contexts in terms of General Accuracy in few-shot settings. This suggests that a targeted approach to context, where only task-relevant information is provided, is generally more beneficial for model performance than a more comprehensive one.

As seen in \autoref{tab:rouge-50percent}, our best performance was obtained with Mistral 7 B on the DocTAET context with a general Accuracy of 89\%, in few-shot and 95\% in zero-shot, markedly outperforming the DocREC context, on the same model which achieves 88\% and 92\% in few-shot and zero-shot respectively. These findings affirm the effectiveness of the DocTAET context not only in structured summary generation but also in its high accuracy for classifying whether a paper includes a leaderboard.

\noindent{\textbf{Performance in Individual Element Extraction (RQ2).}}
The critical nature of precision in language model outputs is particularly pronounced in contexts where inaccuracies carry significant repercussions. Referencing \autoref{tab:f1-50percent} and \ref{tab:precision-50percent}, the extraction of the Task element achieves the highest F1 score, with exact-match evaluations at approximately 33\% and partial-match evaluations around 46\%. The extraction of the Metric element is the second most accurate, with exact-match precision at roughly 24\% and partial-match precision at about 34\%. The Dataset element follows closely, with exact-match precision at around 18\% and partial-match precision at approximately 32.75\%. Although the extraction of the Score element presents challenges, as indicated by single-digit performances in the scores column, it is noteworthy that our context, DocREC, surpasses both DocTAET and DocFULL in performance and sets new state-of-the-art compared to previous approaches in score extractions. This is reflected in the evaluation of both the F1 score and precision metrics, suggesting that DocREC's context is better suited for detailed and accurate element extraction, despite the inherent difficulties of the task.

\noindent{\textbf{Generalizability to Unseen Tasks (RQ3).}}
The ability of language models to generalize from no prior examples, known as zero-shot learning, constitutes a significant aspect of our inquiry (RQ3). This study's empirical evaluation shed light on the models' generalizability when applied to tasks for which they had not been explicitly trained. The analysis, leveraging performance metrics delineated in \autoref{tab:f1-50percent} and \ref{tab:precision-50percent}, suggests that models implemented with DocTAET and DocREC contexts exhibit robust zero-shot learning capabilities. However, it was observed that the DocFULL context yielded suboptimal performance in such zero-shot scenarios. Our best model demonstrated consistent performances across few-shot and zero-shot settings which are imperative for practical applications where encountering previously unseen tasks, datasets, or metrics is a common occurrence

These findings necessitate a nuanced discussion on the implications of context optimization for enhancing a model's generalizability in zeo-shot settings ensuring they remain versatile and effective tools within dynamic research landscapes.  

\section{Conclusions}

In this study, we embarked on a detailed empirical analysis to understand the impact of context selection on the efficacy of large language models (LLMs) in curating leaderboards for scholarly articles. The cornerstone of our research was the evaluation of various context selection techniques and their influence on the precision and reliability of LLMs in generating accurate research paper leaderboards. We conducted thorough experiments to assess the performance of the advanced LLMs, Llama-2 7B and Mistral 7B, leveraging diverse tasks, datasets, and performance metrics.

%
%
%
\bibliographystyle{splncs04}
\bibliography{main}

\appendix

\section{Instructions: Qualitative Examples}
\label{app:instructions}

In this section, we elicit each of the instructions that were considered in this work as formulated in the FLAN 2022 Collection for the SQuAD\_v2 and DROP datasets.

\begin{table}[!h]
\centering
\resizebox{\textwidth}{!}{%
\begin{tabular}{@{}cll@{}}
\toprule
\textbf{ID} & \textbf{SQuAD\_v2 Instructions}                                      & \textbf{DROP Instructions}                                \\ \midrule
1  & Please answer a question about this article. If unanswerable, say "unanswerable". & Answer based on context.                                 \\
2  & \{Context\} \{Question\} If unanswerable, say "unanswerable".                     & Answer this question based on the article.               \\
3  & Try to answer this question if possible (otherwise reply "unanswerable").         & \{Context\} \{Question\}                                 \\
4  & Please answer a question about this article, or say "unanswerable" if not possible.& Answer this question: \{Question\}                       \\
5  & If possible to answer this question, do so (else, reply "unanswerable").          & Read this article and answer this question.              \\
6  & Answer this question, if possible (if impossible, reply "unanswerable").          & Based on the above article, answer a question.           \\
7  & What is the answer? (If it cannot be answered, return "unanswerable").            & Context: \{Context\} Question: \{Question\} Answer:      \\
8  & Now answer this question, if there is an answer (else, "unanswerable").           &                                                          \\ \bottomrule
\end{tabular}%
}
\caption{Comparative Instructions for the SQuAD\_v2 and DROP datasets.}
\label{table:datasets_instructions}
\end{table}
\vspace{-35pt}
\section{ROUGE Evaluation Metrics}
\label{app:rouge}

The ROUGE metrics are commonly used for evaluating the quality of text summarization systems. ROUGE-1 measures the overlap of unigram (single word) units between the generated summary and the reference summary. ROUGE-2 extends this to measure the overlap of bigram (two consecutive word) units. ROUGE-L calculates the longest common subsequence between the generated and reference summaries, which takes into account the order of words. ROUGE-LSum is an extension of ROUGE-L that considers multiple reference summaries by treating them as a single summary. 

\section{Additional Data statistics and Hyperparameters}
\label{app:hyp}

\vspace{-20pt}

\begin{table*}[!h]
\begin{center}
\begin{threeparttable}
\begin{minipage}{\textwidth}
\begin{tabular*}{\textwidth}{@{\extracolsep{\fill}}l|c|c|c@{\extracolsep{\fill}}}
\cmidrule{1-4}%
& \multicolumn{3}{@{}c@{}}{\textbf{Our Corpus}} \\\cmidrule{2-4}%
 & Train &Test-Few-shot & Test Zero-shot \\
\midrule
Papers w/ leaderboards & 7,744/7,987/7025& 722/753/661& 239/241/242  \\
Papers w/o leaderboards &  4,063/4,401/3033 & 604/648/444 & 507/548/353  \\
Total TDM-triples & 402,409/415,788/415,788 & 33,863/34,799/31,213 & 14,604/14,800/14,273  \\
Distinct TDM-triples & 11,814/11,998/11,016 & 1,875/1,917/1,768 & 1,246/1,267/1,219  \\
Distinct \textit{Tasks}       & 1,365/1,374/1,232 & 308/322/292 & 236/236/235  \\
Distinct \textit{Datasets}    & 4,733/4,816/4,473 & 921/947/870 & 642/647/616  \\
Distinct \textit{Metrics}     & 2,845/2,876/2687 & 643/654/603 & 398/412/386  \\
Avg. no. of TDM per paper & 5.13/5.12/5.07 & 4.87/4.81/4.90 & 6.05/6.11/5.88  \\
Avg. no. of TDMS per paper & 6.94/6.95/6.83 & 5.91/5.81/5.93 & 7.81/7.86/7.47\\
\end{tabular*}
\caption{DocFULL/DocREC/DocTAET corpora statistics. The ``papers w/o leaderboard'' refers to papers that do not report leaderboard.}
\label{table:All_Stats}
\end{minipage}
\end{threeparttable}
\end{center}
\end{table*}
\vspace{-30pt}

We used a context length of 2400 and based on GPU availability, a batch size of 2 and gradient\_accumulation\_steps of 4 were used, leading to a final batch size of 8. All experiments were run on five epochs and we used AdafactorSchedule and Adafactor optimizer with scale\_parameter=True, relative\_step=True, warmup\_init=True, lr=1e-4.



\end{document}